\documentclass{article}
\pdfoutput=1

     \usepackage[preprint,nonatbib]{neurips_2020}

\usepackage[utf8]{inputenc} 
\usepackage[T1]{fontenc}    
\usepackage{hyperref}       
\usepackage{url}            
\usepackage{booktabs}       
\usepackage{amsfonts}       
\usepackage{nicefrac}       
\usepackage{microtype}      
\usepackage[normalem]{ulem}

\usepackage{amsmath}
\usepackage{xcolor}
\usepackage{algpseudocode}
\usepackage{algorithm}
\usepackage{siunitx}
\usepackage{graphicx}
\usepackage{multirow}

\newcommand{\etal}{\textit{et al}. }

\title{Learning a Discriminant Latent Space with \\Neural Discriminant Analysis}

\author{%
  Mai Lan Ha \\
  Department of Computer Science\\
  University of Siegen\\
  Germany \\
  \texttt{hamailan@informatik.uni-siegen.de} \\
  \And
  Gianni Franchi \\
  U2IS, ENSTA Paris\\
  Institut Polytechnique de Paris\\
  France\\
  \texttt{gianni.franchi@ensta-paris.fr} \\
  \And
  Emanuel Aldea \\
  Laboratoire SATIE \\
  Paris-Saclay University \\
  France\\
  \texttt{emanuel.aldea@u-psud.fr} \\
  \And
  Volker Blanz \\
  Department of Computer Science\\
  University of Siegen\\
  Germany \\
  \texttt{blanz@informatik.uni-siegen.de} \\
}

\begin{document}

\maketitle

\begin{abstract}

Discriminative features play an important role in image and object classification and also in other fields of research such as semi-supervised learning, fine-grained classification, out of distribution detection. Inspired by Linear Discriminant Analysis (LDA), we propose an optimization called Neural Discriminant Analysis (NDA) for Deep Convolutional Neural Networks (DCNNs). NDA transforms deep features to become more discriminative and, therefore, improves the performances in various tasks. Our proposed optimization has two primary goals for inter- and intra-class variances. The first one is to minimize variances within each individual class. The second goal is to maximize pairwise distances between features coming from different classes. We evaluate our NDA optimization in different research fields: general supervised classification, fine-grained classification, semi-supervised learning, and out of distribution detection. We achieve performance improvements in all the fields compared to baseline methods that do not use NDA. Besides, using NDA, we also surpass the state of the art on the four tasks on various testing datasets.

\end{abstract}

\section{Introduction}

Deep Convolutional Neural Networks (DCNNs) have become standard tools in Computer Vision to classify images with impressive results on ImageNet \cite{imagenet_cvpr09}.
However, having a large-scale annotated dataset like ImageNet is expensive. Therefore, many research techniques such as active learning, semi-supervised learning, or self-supervised learning focus on utilizing small sets of annotated data. 

The discriminative power of latent features is an important factor that affects the performance of classification. In this work, we propose an optimization technique to strengthen the discriminative potential of latent spaces in classification DCNNs. Our approach is based on the objectives of Linear Discriminant Analysis (LDA) which is to minimize the intra-class variance and maximize the inter-class variance. We incorporate these criteria in DCNN training via different losses that can be viewed as constraints for optimizing discriminant features. In other words, our goal is to optimize discriminant analysis for latent features in DCNNs that leads to better classification results, hence the name Neural Discriminant Analysis (NDA). This optimization focuses on solving these three criteria: (i) reducing intra-class variance by minimizing the total distance between features of objects in the same class to their means, (ii) increasing inter-class variance by transforming the features into other feature spaces such that the feature distances between two classes in the target space are pushed further apart, (iii) the previous two criteria should also improve classification accuracy.


The relevance of our approach is shown through the performance improvements of our proposed optimization in various fields such as general classification, semi-supervised learning, out of distribution detection and fine-grained classification.

Semi-supervise learning (SSL), as the name suggests, is a learning technique that combines both labeled and unlabeled data in the training process. The majority of work in this field \cite{sohn2020fixmatch,lee2013pseudo,xie2019unsupervised,xie2019self} uses pseudo labeling and consistency regularization for unsupervised training part. Consistency regularization on unlabeled data is applied on different augmentations of the same input image. One example of the state-of-the-art for unsupervised data augmentation is proposed in \cite{xie2019unsupervised} (UDA). Using a similar approach as UDA, we add our NDA losses to improve the discriminant of both labeled and pseudo labeled data. It shows that NDA improves the final classification result.

One important aspect of a real-world classification system is the ability to identify anomalous data or samples that are significantly different from the rest of the training data. These samples should be detected as out of distribution (OOD) data. This problem has been studied in various papers \cite{blundell2015weight,gal2016dropout,maddox2019simple,franchi2019tradi,lakshminarayanan2017simple,liang2017enhancing}. It turns out that having a more discriminative latent space can also help to improve OOD data detection, and our NDA technique has proved to be useful in this field.


Another specific field in classification is Fine-Grained Visual Classification (FGVC).  The task of FGVC is to classify subordinate classes under one common super-class. Examples are recognizing different breeds of dogs and cats \cite{Khosla_Stanford_Dogs_FGVC2011,parkhi_oxford_pets_2012},  sub-species of birds \cite{WahCUB_200_2011, VanHorn_nabirds_2015}, different models and manufactures of cars and airplanes \cite{Krause_Stanford_Cars_2013,Yang_cars_2015,Maji_airplanes_2013}, sub-types of flowers \cite{Nilsback_flowers_2008} or natural species \cite{Horn_iNat_2018} and so on. On the one hand, it is challenging because the subordinate classes share the same visual structures and appearances; the differences are very subtle. In many cases, it requires domain experts to distinguish and label these classes by recognizing their discriminative features on specific parts of the objects. Therefore, it is also a great challenge to obtain large-scale datasets for FGVC. On the other hand, the intra-class variance can be visually higher than the inter-class variance. Such cases can be seen in different colors and poses of objects in the same class. Our proposed NDA directly addresses the challenge of FGVC, and our experiments show that we achieve improvements on various FGVC dataset using NDA optimization.

What makes our method interesting is not only the improvement of performance, but also the fact that NDA can be deployed easily  as a component to an existing network for different tasks. We can connect NDA to any layer of any existing DCNN and make it end-to-end. However, in this paper, we just introduce NDA on the pre logit layer, which contains have high-level information that is useful for classification. Our contribution is to propose a method to learn a discriminant latent space for DCNNs. We conduct various experiments to show the applicability of NDA in different research topics. Our proposed optimization helps to improve fine-grained classification and semi-supervised learning performance. In addition, in OOD data detection, our algorithm helps in obtaining a confidence score that is better calibrated and more effective for detecting OOD data.

\section{Related Work} 

\textbf{Discriminant Analysis with DCNN:} To improve the classification performance, one option is to use discriminant analysis incorporated with DCNNs. Mao \etal \cite{Mao_1993_nonlinear_analysis} proposed a nonlinear discriminant analysis network while others use linear discriminant analysis \cite{Wang_ijcai2017_2D_LDA, Dorfer_ICLR16_deep_LDA, Zhong_CDA_ICPR_2018, Li_ciss19_da}. The common idea of these methods is to implement the discriminative principles of LDA in trainings of DCNNs with different implementations. Zhong \etal \cite{Zhong_CDA_ICPR_2018} proposed a method to optimize feature vectors with respect to the centers of all the classes. The training is done in single branch of a DCNN. Dorfer \etal \cite{Dorfer_ICLR16_deep_LDA} compute eigenvalues on each training batch to minimize between class variances. However, this method is slow due to the eigenvalue computations. Our proposed method is based on a Siamese training without the need for solving the eigen problem. A preliminary work building on this general strategy and addressing only FGVC has been accepted in ICIP 2020 \cite{annomynous_nda}. Here, we generalize and extend the idea of NDA in a combined loss function and a joint training procedure, and we demonstrate the general usefulness of this new framework in a wider range of applications. 

\textbf{Semi-supervised learning (SSL):} Training DCNNs often needs a large amount of data, which can be expensive to annotate. Semi-supervised learning (SSL) techniques overcome this challenge by using an auxiliary dataset together with a small annotated target dataset. Auxiliary datasets can come from different datasets such as \cite{dvornik2019diversity,gidaris2019boosting}, or are formed by a subset of the main training dataset of which the annotations are omitted \cite{sohn2020fixmatch,lee2013pseudo,xie2019unsupervised,xie2019self}. We consider the latter case since the unlabeled data is easily obtained from the main labeled dataset. A set of methods \cite{sohn2020fixmatch,lee2013pseudo,xie2019unsupervised} train DCNNs on the unlabeled dataset by applying two random data transformations on the unlabeled data and forcing the DCNNs to be consistent between the two predictions on the two transformed images.

\textbf{Out of distribution (OOD) data detection:} OOD detection is challenging since it involves understanding the limitations of a trained DCNN. Relying on an uncertainty measure is a solution for this task since the data with high uncertainty can be considered as OOD. Uncertainty in deep learning has been initiated with Bayesian DCNNs, which assume a posterior distribution on the weights given the dataset. It can be estimated directly \cite{blundell2015weight} or indirectly via dropout \cite{gal2016dropout}. Thanks to the marginalization principle, the results are integrated over the posterior distribution. Among different Bayesian DCNNs, MC Dropout \cite{gal2016dropout} allows a fast estimation of the uncertainty, while SWAG \cite{maddox2019simple} and TRADI \cite{franchi2019tradi}  estimate the posterior distribution by studying the training process of the DCNNs. Uncertainty has also been studied via ensemble methods on Deep Ensembles \cite{lakshminarayanan2017simple} and OVNNI \cite{franchi2020OVNNI} which presents state-of-the-art results.  Finally, other methods try to detect OOD by learning the loss of the DCNN  such as Confidnet \cite{corbiere2019addressing}, or by proposing algorithms specific for this task such as ODIN \cite{liang2017enhancing} that is tuned with OOD data.

\textbf{Fine-grained visual classification (FGVC):} The principles of LDA address directly to the challenges that FGVC faces. In the field of FGVC, the inter-class differences are often subtle. Experts distinguish subordinate classes based on specific parts of the objects. Therefore, a straight-forward approach is to learn features of object parts \cite{Farrell_2011,Khosla_Stanford_Dogs_FGVC2011,Parkhi_cats_dogs_2011,Liu_dog_part_2012,branson2014bird,ZhangECCV14,ZhangCVPR14,Krause_parts_2014,Lin_deep_LAC_2015,ZhangSGD15,Huang_2016_CVPR,Zhang_SPDA_CNN_2016}. This approach requires heavy part annotations from domain experts, and therefore it is difficult to extend to larger scale datasets. Some other works rely on attribute annotations and text descriptors \cite{Vedaldi_detail_attribute_CVPR_2014,Reed_visual_descriptions_CVPR_2016,He_vision_language_CVPR_2017,Xu_semantic_embedding_IJCAI_2018}. Stepping away from those types of annotations, another set of works focus on learning and using visual attention on discriminative regions \cite{Xiao_attention_2015,Liu_FullyCA_2016,Zheng_multi_attention_2017,Zhao_attention_2017,Fu_look_closer_2017,Peng_object_part_attention_TIP_2018,Yang_learn_navigate_2018,Sun_multi_attention_2018}. Analyzing the filter responses from DCNNs has also led to good part descriptors and localization \cite{Wang_granular_descriptor_2015,Zhang_picking_resposes_2016}. Utilizing the internal responses from DCNNs, different pooling techniques have also been developed such as bilinear pooling to study the interactions of sets of local features \cite{Lin_bilinear_pooling_2015,Gao_compact_pooling_2016,Lin_improve_pooling_2017,Cui_kernel_pooling_2017,Cai_high_order_hierarchical_2017,Yu_hierarchical_pooling_2018,Wei_Grassmann_pooling_2018}. Besides the high intra-class and low inter-class variance challenge, FGVC also faces a problem from small scale datasets. In order to address this issue, researchers work on different strategies to collect more relevant images to enrich the datasets \cite{Xie_myperclass_augment_CVPR_2015,Krause_noisy_data_ECCV_2016,Gebru_in_the_wild_ICCV_2017,Zhang_auxiliary_data_ECCV_2018,Xu_webly_supervised_PAMI_2018,Cui_iNatTransfer_CVPR_2018} or to employ human in the loop and human interaction to bootstrap datasets \cite{Branson_human_in_loop_ECCV_2010,Cui_bootstrap_CVPR_2016,Deng_crowd_wisdom_PAMI_2016}.

\section{Neural Discriminant Analysis (NDA)}

Features for image classifications can be obtained from various types of network models. A straightforward approach is to extract features from pre-trained models on the ImageNet dataset \cite{imagenet_cvpr09} with 1,000 classes such as Inception \cite{Szegedy_Rethinking_Inception_CVPR_2016}, ResNet \cite{He_ResNet_CVPR_2016,He_ResNet_ECCV_2016}, etc. However, they are not the most discriminative features for various tasks such as fine-grained visual classification (FGVC), semi-supervised learning (SSL), or out of distribution (OOD) detection. Hence, we opt to improve their discriminative potential by learning a discriminant analysis on these deep feature spaces.

\subsection{Linear Discriminant Analysis (LDA)}

Let $f$ be a Deep Convolutional Neural Network (DCNN) with a set of weights $ \omega$. An input image is denoted as $x$. Hence the classification result of the DCNN $f$ applied on the input image $x$ is $f(x,\omega)$. In addition we denote $f^l(x,\omega)$ the latent space of the DCNN corresponding to the features extracted at layer $l$ which is usually located before the classification layer. 

Let $\mathcal{X}_{\mbox{train}}=\{x_k,y_k\}_{k\in [1,N]}$ be a training set composed of $N$ data samples, and 
$x_k$ is an input that can be a signal or an image, while $y_k$ is its corresponding class. We assume that the dataset contains $K$ classes. Let $\mu_j$ be the empirical mean of each class $j^{th}$ that has $N_j$ data samples, and $\mu$ the empirical global mean. The within scatter matrix $S_W$ and the between scatter matrix $S_B$ are computed as follows:
\vspace{-5pt}
\begin{equation}
\small
    S_B = \frac{1}{N}\sum_{j=1}^K N_j( \mu_j -\mu) (\mu_j -\mu) ^T,\quad
    S_W = \frac{1}{N}\sum_{j=1}^K \sum_{i=1}^{N_j}(x_i.\delta_{y_i=j} -\mu_j) (x_i.\delta_{y_i=j} -\mu_j)^T
\end{equation}


where $\delta_{y_i=j} $ is the Dirac function equal to $0$ when the class of $x_i$ is different to $j$ and $1$ otherwise.

The objective of the LDA is to learn a projection $U\in M_{D,d}(\mathbb{R}$) that maps the data from the initial dimension $D$ to a smaller dimension $d$ such that the inter-class variance is maximized and the intra-class variance is minimized. The projection matrix U needs to satisfy Fisher’s condition:
\vspace{-3pt}
\begin{eqnarray}\label{S_W}
    U_{lda} = \underset{U}{\arg\max} \frac{|U^T S_B U|}{|U^T S_W U|}
\end{eqnarray}


\subsection{From Linear Discriminant Analysis to Neural Discriminant Analysis} \label{subsection:discriminant_analysis}

\begin{figure}[tp!]
    \centering
    \includegraphics[width=0.85\linewidth]{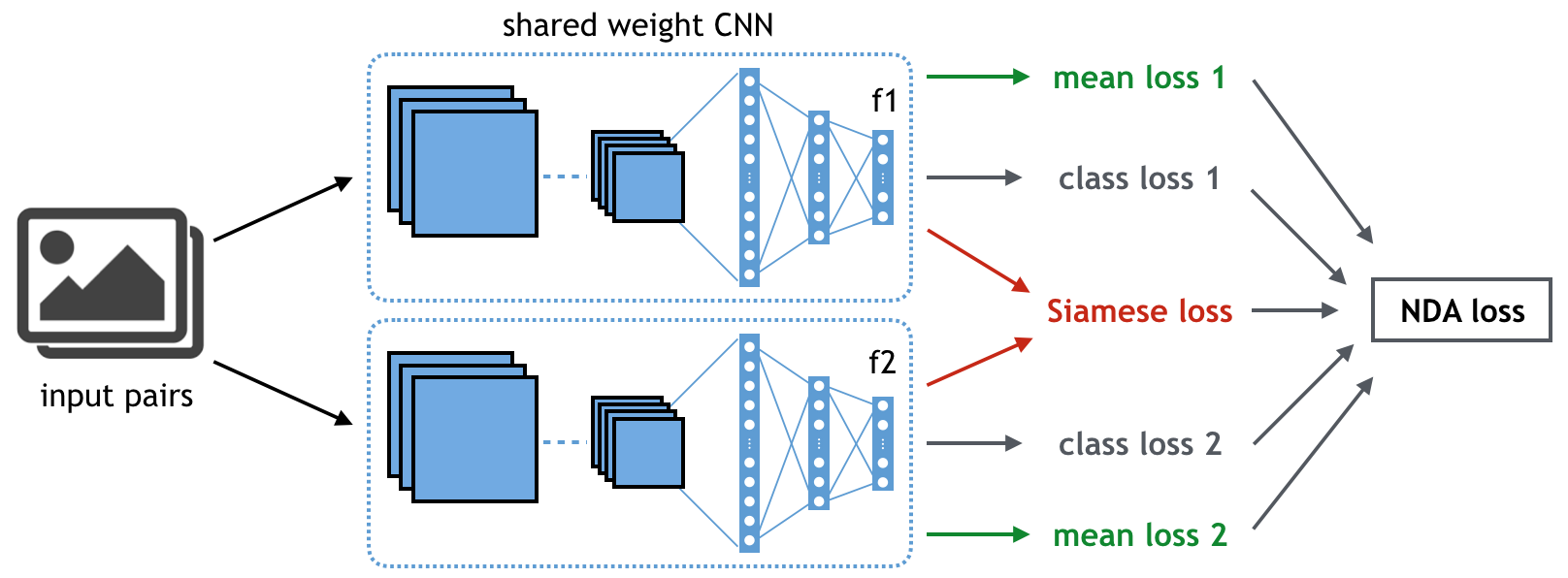}
    \caption{A demonstration of NDA optimization. The NDA loss is the weighted sum of two Classification losses, two Mean losses and a Siamese loss produced by a Siamese network.}
    \label{fig:NDA_overview}
\end{figure}

Inspired by the objectives of LDA, we propose an optimization to learn a discriminative latent space. It minimizes the intra-class variance and maximizes the inter-class variance of the latent space. The optimization's objectives for $f$ are the following:

\begin{itemize}
    \item \textbf{Maximizing the classification results:}  Let the classification results be $q(x_k)=f(x_k,\omega)$. Maximizing the results is equivalent to minimizing the categorical cross entropy loss:
    
    \begin{equation} \label{equation:category_cross_entropy}
        \mathcal{L}_{class} = -\sum_{\forall x_k}p(x_k)log(q(x_k)),
    \end{equation}
    
    where $p(x_k)$ is the one-hot encoded classification ground-truth of the target $y_k$. 
    
    \item \textbf{Minimizing intra-class variance:} This loss minimizes the total distances for all the feature points $f^l(x_{i},\omega)$ to their respective class mean feature, and is calculated as below:
    \begin{equation} \label{equation:mean_variance}
        \mathcal{L}_{Mean} = \frac{1}{N}\sum_{j=1}^{K}\sum_{i=1}^N D_W( f^l(x_{i} ,\omega ), \mu_j^l).\delta_{y_{i}=j}
    \end{equation}
    where $\mu_j^l$ is the mean of latent features of class $j$, which is calculated at the beginning of each epoch. $D_W$ is a distance function or a dissimilarity measure. The mean loss can be either the exact equation \ref{equation:mean_variance} where $D_W$ is the L2-norm, or we can also use the prototypical loss \cite{snell2017prototypical} where the prototypes are the mean features. Therefore, instead of just evaluating the mean distance we compute the softmax of the mean distances, followed by cross-entropy.
    
    \item \textbf{Maximizing inter-class variance:} With this optimization, we propose to use pairs of images $(x_{i1},x_{i2})$. If this pair of images belongs to the same class, we want to reduce the distance between the image latent space, otherwise, we want to increase the distance between them. The effect is to push features from different classes apart from each other while keeping features within the same class close to each other. Let $y_{i1i2}=0$ if the two images $(x_{i1}$ and $x_{i2})$  are from the same class and $y_{i1i2}=1$ if they are from different classes. The optimization for inter-class variance minimizes the following function:
    \begin{align} \label{equation:inter_class_variance}
        \mathcal{L}_{Siamese} &= \frac{1}{N}\sum_{i1=1}^N\sum_{i2=1}^N ((1-y_{i1i2}) D_S(f^l(x_{i1},\omega),  f^l(x_{i2},\omega )) - y_{i1i2} D_S( f^l(x_{i1},\omega ) ,f^l(x_{i2},\omega))
    \end{align}
    where $D_S$ is a distance function, which is L2-norm in all experiments.
    \item \textbf{Total loss:} The core structure of combining losses for NDA optimization is a Siamese architecture. A pair of input images are passed through a shared weight Siamese network. The Siamese network is designed with 5 losses: two Classification losses, two Mean losses (one Classification loss and one Mean loss for each image), and a Siamese loss (Fig~\ref{fig:NDA_overview}). The Classification loss is the categorical cross entropy loss defined in Equation~\ref{equation:category_cross_entropy}. The Mean loss is defined as in Equation~\ref{equation:mean_variance} and the Siamese loss is defined as in Equation~\ref{equation:inter_class_variance}. The combination loss is defined as the weighted sum of all the losses as follows:

    \begin{equation} \label{equation:total_loss}
        \mathcal{L}_{NDA} = \alpha(\mathcal{L}_{Class1} + \mathcal{L}_{Class2}) + \beta(\mathcal{L}_{Mean1} + \mathcal{L}_{Mean2} + \gamma*\mathcal{L}_{Siamese}) 
    \end{equation}

    where $\mathcal{L}_{Class1}$ and $\mathcal{L}_{Mean1}$ are the Classification loss and Mean loss for the first input image. $\mathcal{L}_{Class2}$ and $\mathcal{L}_{Mean2}$ are the Classification loss and Mean loss for the second input image. The coefficients $\alpha$, $\beta$, and $\gamma$ are fine-tuned hyper-parameters.
\end{itemize}

The NDA model (Figure~\ref{fig:NDA_overview}) is designed as follows: the images of a batch are provided to a DCNN that produces the latent space representation of these images and the classification output target. To optimize jointly the classification and inter-class and intra-class variance of the latent space, we first calculate the mean feature for every class at the beginning of an epoch. Then, we provide two batches of images to the DCNN. Each batch is used for the Classification loss and the Mean loss. The combination of two batches is used for the Siamese loss. We then combine all the losses as in Eq~\ref{equation:total_loss}.


    
    
    

\section{Experiments}

We run the experiments with different values for the hyper-parameters $\alpha$, $\beta$, and $\gamma$. While we cannot extensively search for many configurations of the hyper-parameters due to limited resources, we find that the configuration $\alpha = 1$, $\beta = 1e-3$, and $\gamma = 1$ provides slightly better results. For the experiments in SSL and OOD, we implement the prototypical loss \cite{snell2017prototypical} on the Mean loss. Instead of just using L2 distance on the mean class vector, we use linear distance and normalize the distance by a softmax function. We will briefly describe experiments for different applications and their results in this Section. 

\subsection{General Supervised Classification}

We train end-to-end networks for classification on the CIFAR-10 dataset \cite{krizhevsky2009learning}. We use the official (train, test) split of the dataset. However, we further split the training set to 80\% for training data and 20\% for validation. We only validate our training on the validation set and test on the test set. We use two different network architectures for the Siamese: AlexNet \cite{Krizhevsky_NIPS2012_AlexNet} with Kaggle implementation and ResNet50  \cite{He_ResNet_ECCV_2016}. The results for AlexNet Kaggle are shown in Table~\ref{table:nda_cda_cifar10} and ResNet50 are in Table~\ref{table:nda_resnet_cifar10}. 

\begin{table}[h!]
    \centering
    \scalebox{0.9}
    {
    \begin{tabular}{c|c|c|c}
    \hline
        & Baseline & Optimization & Improvement Over Baseline\\
    \hline
        CDA \cite{Zhong_CDA_ICPR_2018} & 60.2\% & 62.5\% & 2.3\% \\
        NDA (ours) & 64.7\% & \textbf{70.9\%} & \textbf{6.2\%} \\
    \hline
    \end{tabular}
    } 
    \caption{Comparison between NDA and CDA \cite{Zhong_CDA_ICPR_2018} on CIFAR-10 dataset using AlexNet Kaggle \cite{Krizhevsky_NIPS2012_AlexNet}. NDA has significant better result than CDA. The NDA result is averaged over five runs.}
    \label{table:nda_cda_cifar10}
\end{table}
\vspace{-5pt}
We compare with a competing method "Convolutional Discriminant Analysis" (CDA) proposed by Zhong \etal \cite{Zhong_CDA_ICPR_2018} using AlexNet Kaggle base network. Different from our Siamese architecture, CDA uses a single branch CNN with a single input image. With the Siamese network, we can specify the Mean loss and the Siamese loss explicitly. On the contrary, CDA's objective is that an input image trained to be close to its class center and further away from other classes' centers. NDA achieves 70.9\% accuracy whereas CDA only reaches to 62.5\%. Compared to the baseline, CDA has 2.3\% increase and NDA has 6.2\% improvement. We re-implement AlexNet Kaggle using Keras-Tensorflow. Due to different framework supports, our baseline has higher accuracy than the CDA's. The numerical results conclude that our NDA achieves significantly better results than CDA and much higher improvement from the baseline.

We also test our proposed NDA using ResNet50 \cite{He_ResNet_ECCV_2016}. We use the code for the model provided by Keras \footnote{ \url{https://keras.io/examples/cifar10_resnet/}}, as well as the pre-defined learning schedule and 200 epochs for each training. It is reported that the accuracy of ResNet50 on CIFAR-10 is 93.0\%. However it is important to take note that this accuracy is reported for training that validates on the test set's accuracy. On the contrary, we split the default training data into 80\% for training and 20\% for validation to avoid over-fitting on the test data. This also results in having less data for training. We achieve a baseline accuracy of 91.8\%, while NDA improves the accuracy to 93.0\%. We also experiment with using the prototypical loss \cite{snell2017prototypical} to optimize for the Mean loss. This setup improves the result to 93.3\%, surpasses the state-of-the-art results. The NDA results are averaged over five runs. 

\begin{table}[h!]
    \centering
    \scalebox{0.9}
    {
    \begin{tabular}{c|c|c|c|c}
    \hline
        NetCCE \cite{Dorfer_ICLR16_deep_LDA} & SOTA (*) & Baseline & NDA & NDA + prototypical loss \cite{snell2017prototypical}\\
    \hline
        92.9\% & 93.0\% & 91.8\% & \textbf{93.0\%} & \textbf{93.3\%} \\
    \hline
    \end{tabular}
    } 
    \caption{Comparison between NDA and the state-of-the-art result on CIFAR-10 with ResNet50. (*) The SOTA (state of the art) result is validated on the test set accuracy. For all the NDA training, we train and validate on (80\%, 20\%) split on the official training set and the results are reported on the test set. This is the same for NetCCE method, except that NetCCE uses VGG network. NDA's results surpasses both NetCCE and the SOTA results. The NDA results are averaged over five runs.}
    \label{table:nda_resnet_cifar10}
\end{table}

We also evaluate NDA optimization on CIFAR-100 dataset using WideResNet 28x10 network\cite{zagoruyko2016wide}. The accuracy of the network without NDA is 73.9\% and 76.4\% with NDA. Our optimization increases the baseline result by 2.5\%.



\subsection{Semi-Supervised Learning (SSL) Classification}

In this section, we propose to use NDA optimization in a semi-supervised learning (SSL) task. We first randomly select 250 images with labels from CIFAR-10 dataset, 25 images per class for 10 classes. These 250 images are used in supervised training for classification. We also select 2,000 labeled images from the official training set as validation data. The rest of the data in the training set is used as unlabeled data. We use WideResNet 28x10 architecture \cite{zagoruyko2016wide} and consistency loss, similar to those used in UDA \cite{xie2019unsupervised}, for training SSL.

The training is divided into two phases. In the first phase, we have two losses: one for supervised training and one for unsupervised training. The first loss is the categorical cross-entropy which is used for supervised training on the 250 labeled images. We also use the unlabeled set to train the DCNN with a consistency loss. The consistency loss is the Kullback–Leibler (KL) divergence on the class probability outputs of two transformations of an input image. The first transformation is a random horizontal flip and random cropping, the second one is the more complex transformation proposed by UDA \cite{xie2019unsupervised} that uses RandAugment \cite{Cubuk2019_RandAugment}. The consistency loss makes sure that the class probability distributions of two transformations from the same image are similar.



In the second phase, we use a deep ensemble with three models trained in the first phase to pseudo label the unlabeled set. The pseudo labels are created by averaging the predictions from all three models. We also use the deep ensemble networks to compute the confidence score of an unlabeled image, which is the max probability class (MCP) of the average result. If the confidence score is higher than $0.95$, we will select the image and its pseudo label for training. We combine the pseudo labeled data and 250 labeled images to continue training the three ensemble models for classification using cross-entropy loss. After every training epoch, if the performance of a deep ensemble network is improved in the second phase, we will use it to update its predecessor network in the first phase. Even though the technique is similar to UDA \cite{xie2019unsupervised}, the deep ensemble pseudo labeling is a new contribution.

The result of our SSL technique on CIFAR-10 is 90.0\%. When incorporating NDA optimization into the loss in the second phase, we achieve an accuracy of 90.5\%, an increase of 0.5\% over the baseline. 

\subsection{Out Of Distribution Detection (OOD)}

In this subsection, we show the performance of NDA for detecting OOD data. We train a Resnet50 architecture on CIFAR-10 and test it on a combination of images from CIFAR-10  and SVHN, which is a commonly used dataset for OOD. The goal of this section is to show that without any hard extension, NDA can reach a performance close to the state of the art results.

For ResNet50 trained on CIFAR-10, the Maximum Class Probability (MCP), which is the confidence score \cite{hendrycks2016baseline} for a classical training, seems to be close to 0.9 for most of the OOD data. This behaviour is wrong and could lead to a wrong use of DCNN. Furthermore, for the NDA training we also consider the MCP as a confidence score, yet in this case the OOD data have a lower score.

To have a better quantitative evaluation, we compare NDA with most competitive algorithms and use the criteria metrics proposed in \cite{hendrycks2016baseline}. We evaluate our method and compare with existing work using AUC, AUPR and FPR-95\%-TPR metrics. We also evaluate the Expected Calibration Error (ECE) \cite{guo2017calibration}, which checks if the confidence scores are reliable. NDA has better results than the Bayesian DCNN and the learning loss techniques. The only one that surpasses NDA is Deep Ensembles that needs to train several DCNNs, thus, requires more training time. We also compare NDA to ODIN. It shows that NDA is comparable to strategies that fine-tune the DCNNs using OOD data.

 \begin{table}[htbp]
 \begin{center}
 \scalebox{0.9}
 {
 \begin{tabular}{|l|l|l|c|c|c|c|}
 \hline
 Dataset         & Model           & OOD technique  & AUC  & AUPR & FPR-95\%-TPR  & ECE \\ \hline
                 &                 & Baseline (MCP) & 94.0 & 96.0 & 24.6 & 0.606        \\ \cline{3-7} 
                 &                 & MC Dropout      & 80.4 & 89.7 & 61.5 & 0.606       \\ \cline{3-7} 
                 &                 & Deep Ensemble  & \textbf{93.0} & \textbf{96.2} & \textbf{19.3}  & 0.422        \\ \cline{3-7} 
 CIFAR10         & ResNet50        & ODIN           & 80.3 & 89.9 & 61.3  & 0.606       \\ \cline{3-7} 
                 &                 & ConfidNET      & 84.8 & 94.0  & 68.3 & 0.706        \\ \cline{3-7} 
                 &                 & OVNNI   & \textbf{92.2} & \textbf{95.8} & 23.3   &\textbf{ 0.187}  \\ \cline{3-7} 
            &                 & NDA   & \textbf{91.7} & \textbf{95.4} & 24.6   & 0.538 \\  \hline

 \end{tabular}
 } 
  \caption{Comparative results obtained on the OOD task. }\label{table:OODresults2_ugly}
 \end{center}
 \end{table}
\vspace{-25pt}

\subsection{Fine-grained Visual Classification}

We evaluate our proposed method on the following FGVC datasets: Stanford-Dogs \cite{Khosla_Stanford_Dogs_FGVC2011} that contains 120 breeds of dogs, CUB-200-2011 \cite{WahCUB_200_2011} that has 200 types of birds, Flower-102 \cite{Nilsback_flowers_2008} with 102 types of flowers, Stanford-Cars \cite{Yang_cars_2015} that has 196 types of cars and NABirds \cite{VanHorn_nabirds_2015} with 555 classes of birds. We use transfer learning Inception-ResNet-V2, Inception-ResNet-V2-SE and Inception-V3-iNat models from Cui \etal \cite{Cui_iNatTransfer_CVPR_2018} as base networks to compare across all datasets. In this experiment, we alternate the Mean loss and the Siamese loss in the optimization. 

\begin{table}[h!]
    \centering
    \scalebox{0.9}
    {
    \begin{tabular}{c|c|c|c|c|c}
    \hline
    \textbf{Method} & \textbf{CUB-200} & \textbf{Stanford} & \textbf{Stanford} & \textbf{Flower} & \textbf{NABirds}  \\
     &  & \textbf{Dogs} & \textbf{Cars} & \textbf{102} &  \\
    \hline
    Inception-V3  from \cite{Cui_iNatTransfer_CVPR_2018} & 82.8 - 89.3 & 78.5 - 85.2 & 88.3 - 91.4 & 96.3 - 97.7 & 82.0 - 87.9 \\
    Best of Transfer Learning \cite{Cui_iNatTransfer_CVPR_2018} & 89.6 & 88.0 & 93.5 & \textbf{97.7} & 87.9 \\
    NDA (Inception-V3-iNat) (ours) & 87.4 & 89.1 & \textbf{99.9} & 95.5 & 83.9 \\
    NDA (Inc-Res-V2) (ours) & \textbf{90.1} & 95.3 & 97.4 & \textbf{97.7} & 88.4 \\ 
    NDA (Inc-Res-V2-SE) (ours) & 89.7 & \textbf{95.5} & \textbf{99.9} & \textbf{97.7} & \textbf{89.5} \\ 
    \hline
    \end{tabular}
    } 
    \caption{FGVC results. The features used in our methods are from networks in \cite{Cui_iNatTransfer_CVPR_2018}. The best results are reported for Transfer Learning \cite{Cui_iNatTransfer_CVPR_2018} which are from their Inception-V3 or Inception-ResNet-v2-SE networks. Inception-V3 results from \cite{Cui_iNatTransfer_CVPR_2018} are reported as the range of accuracy from all the data sampling strategies as described in \cite{Cui_iNatTransfer_CVPR_2018}. "Inc-Res" is short for Inception-ResNet.}
    \label{table:main_results}
\end{table}
\vspace{-10pt}
We use features extracted from Inception-ResNet-V2, Inception-ResNet-V2-SE and Inception-V3-iNat networks from \cite{Cui_iNatTransfer_CVPR_2018}. It is worth to note that the best performances of transfer learning networks in \cite{Cui_iNatTransfer_CVPR_2018} are from Inception-V3 and Inception-ResNet-V2-SE. Even though the features from Inception-ResNet-V2 do not produce the best results, we are still able to top the state-of-the-art in Stanford-Cars, Flower-102 and NABirds datasets. Authors in \cite{Cui_iNatTransfer_CVPR_2018} trained Inception-V3 by using different data-sampling strategies but there is no strategy that is able to consistently produce the best transfer learning result on all the datasets. It is unclear which of the strategies has been applied in the publicly available network that we use. Thus, for Inception-V3 results from \cite{Cui_iNatTransfer_CVPR_2018}, we report the range of accuracy from all the data sampling strategies. The training is repeated 10 times, with different random initializations. The reported results are the average performance over 10 runs.


With NDA optimization on the Stanford-Cars dataset using features extracted from Inception-ResNet-V2-SE and Inception-V3-iNat, we raise the accuracy to 99.9\% consistently throughout all 10 runs (the standard deviation is therefore 0). Without the NDA optimization, the average accuracy of Inception-ResNet-V2-SE is 97.4\%, and the results fluctuate from 89.1\% to 99.9\% (standard deviation is 3.88). This shows the consistency of the NDA optimization.

\section{Discussion}

\subsection{Learning a Discriminant Latent Space}

We choose to integrate the losses of discriminant analysis in Section~\ref{subsection:discriminant_analysis} into Deep Convolutional Neural Networks and form Neural Discriminant Analysis (NDA) networks due to several reasons. 
First, we can easily add an NDA component to an existing classification model in between the final feature layer and the prediction layer. The NDA approach is flexible in design and transforms the latent space to be more discriminative. 
Secondly, when we train a DCNN to classify images, the first part of the DCNN (such as convolutional layers) learns the features while the last part is specialized to classify. By incorporating NDA  into DCNNs and making it end-to-end networks, it will help to improve the discriminative of features in both latent space and other convolutional layers. Thus, our technique is useful in cases where we have relatively few training data. 


The loss that we propose allows us to have a latent space that is more discriminative. Also, if we consider that $D_S$ is the square norm 2 $\| .. \| ^2$ then the Siamese loss for the data of a batch $B$ that have the same mean turn out to be : 
$\sum_{j=1}^{K}\frac{N_{B,j}}{N}\sum_{i=1}^{N_b}   \| f^l(x_i,\omega ) -\mu^{l,B}_{j}\|^2 1_{y_{i}=j}$, with $\mu^{l,B}_{j}=1/N_{B,j}\sum_{i=1}^{N_b} f^l(x_i,\omega) 1_{y_i=j} $ the empirical mean of on the batch $B$ of the latent space of the class $j$, and  $N_{B,j}$ the number of data on the batch $B$ of class $j$. While the mean loss tries to make the latent space converges to empirical mean of size $N_j$, the intra-class variance part of the Siamese loss tries to make it converge to an empirical mean of size $N_{B,j} \ll N_j$. Hence the empirical mean of the mean loss is more accurate. In addition, the mean loss acts like an anchor that stabilizes the DCNN.

\subsection{Reliability}

We compute and report standard deviations of the accuracy across 10 runs per dataset per network on FGVC task in Table~\ref{table:confident_interval_results}. The standard deviations are consistently lower in all NDA optimization results compared to transfer learning. It shows that the NDA optimization transforms the features such that the classification becomes more stable and reliable. High average with small standard deviation results are much more desired, compared to lower average and high standard deviation.

\begin{table}[!h]
    \centering
    \scalebox{0.9}
    {
    \begin{tabular}{c|c|c|c|c|c|c}
    \hline
    \textbf{Model} & \textbf{Method} & \textbf{CUB} & \textbf{Stanford} & \textbf{Stanford} & \textbf{Flower} & \textbf{NABirds}  \\
     &  & \textbf{200} & \textbf{Dogs} & \textbf{Cars} & \textbf{102} &   \\
    \hline
    Inc-ResNet-V2 &
    TL / NDA & 0.74 / \textbf{0.28} & 0.96 / \textbf{0.81} & 2.53 / \textbf{1.80} &  0.22 / \textbf{0.10} & 0.73 / \textbf{0.08} \\
    \hline
    Inc-ResNet-V2-SE &
    TL / NDA & 0.69 / \textbf{0.17} & 0.37 / \textbf{0.11} & 3.88 / \textbf{0.00} & \textbf{0.13} / \textbf{0.13} & \textbf{0.06} / \textbf{0.06} \\
    \hline
    Inception-V3-iNat &
    TL / NDA & 3.96 / \textbf{0.27} & 0.86 / \textbf{0.40} & 8.32 / \textbf{0.00} & 0.50 / \textbf{0.23} & \textbf{0.09} / 0.10 \\
    \hline
    \end{tabular}
    } 
    \caption{Training stability: standard deviations of accuracy over 10 trainings. The standard deviations for classification using NDA optimization are consistently smaller than those of classification without NDA. It shows that the NDA optimization produces more stable results across different trainings.}
    \label{table:confident_interval_results}
\end{table} 
\vspace{-10pt}


\subsection{NDA versus Siamese}
\begin{table}[htp!]
    \centering
    \scalebox{0.9}
    {
    \begin{tabular}{c|c|c|c|c|c}
    \hline
    \textbf{Method} & \textbf{CUB-200} & \textbf{Stanford-Dogs} & \textbf{Stanford-Cars} & \textbf{Flower-102} & \textbf{NABirds}  \\
    \hline
    Siamese (Inc-Res-V2) & 89.7 & 94.9 & 96.5 & 97.5 & 88.2 \\
    NDA (Inc-Res-V2) (ours) & \textbf{90.1} & \textbf{95.3} & \textbf{97.4} & \textbf{97.7} & \textbf{88.4} \\ 
    \hline
    \end{tabular}
    } 
    \caption{Comparison of results for NDA optimization and Siamese network (without Mean Loss). All the parameters are kept the same throughout the whole paper. All of our results are averaged over 10 training runs. NDA produces better results than using Siamese loss alone in all five datasets.}
    \label{table:NDA_vs_Siamese}
\end{table}
\vspace{-10pt}
We also experiment with an optimization that uses only Siamese Loss alone on FGVC task, the performance drops (Table~\ref{table:NDA_vs_Siamese}). Without the Mean Loss, it lacks a strong and explicit constraint for intra-class optimization. Without Siamese Loss, there is no inter-class optimization.

\section{Conclusion}
Inspired by the objectives of Linear Discriminant Analysis (LDA) and making use of the power of DCNNs, we propose a Neural Discriminant Analysis (NDA) optimization that is useful for many research fields. Our proposed NDA consists of Mean losses, Siamese losses and Classification losses. The combination of all the losses  minimizes the intra-class variance and maximizes the inter-class variance in the deep feature domain. We validate the NDA optimization in four different topics: general supervised classification, semi-supervised learning, out of distribution detection and fine-grained classification. The experiments show that NDA always improves the classification results over the baseline. We also obtain state-of-the-art accuracy on CIFAR-10 and several popular FGVC datasets. NDA results on OOD indicate that it can help to detect OOD data with competitive results to most of the methods, except for Deep Ensembles which requires much more training time. The analysis for FGVC shows that our optimization provides more stable and reliable results. 


\section{Broader Impact}

The discriminant analysis on latent space for features of DCNNs improves the discriminative ability of the networks, especially for cases with less training data such as semi-supervised learning and fine-grained classification. Therefore, it is economically beneficial in helping developing algorithms to achieve good performance on less expensive annotated data. Our work pulls the focus from data-driven approaches that purely depend on millions of annotated images to the direction of incorporating classical machine learning principles in deep learning, using less training data but still achieve excellent results.

We also want to improve the performance in out of distribution detection using NDA optimization. This is an essential feature for real-world applications, especially in medical imaging, to identify anomalous data that the networks have not been trained on or seen before. In those systems, it is crucial not to assign a trained label to anomalous data blindly.


Our proposed method is easier to apply to existing network architectures than \cite{annomynous_nda}, faster training than \cite{Dorfer_ICLR16_deep_LDA}, provides higher accuracy than \cite{Zhong_CDA_ICPR_2018} and is more general than \cite{Li_ciss19_da}. Aspects less often mentioned in published papers are the stability of the training and the reliability of the outcome networks. In this paper, we study the variance of the classification results. It shows that NDA optimization stabilizes the training. It helps to reduce the fluctuations in networks' performance. We would like to emphasize that the reliability in network training is also an important aspect besides the performance itself.





\medskip

\small

\bibliographystyle{unsrt}
\bibliography{main}

\begin{thebibliography}{10}

\bibitem{imagenet_cvpr09}
J.~Deng, W.~Dong, R.~Socher, L.-J. Li, K.~Li, and L.~Fei-Fei.
\newblock {ImageNet: A Large-Scale Hierarchical Image Database}.
\newblock In {\em CVPR09}, 2009.

\bibitem{sohn2020fixmatch}
Kihyuk Sohn, David Berthelot, Chun-Liang Li, Zizhao Zhang, Nicholas Carlini,
  Ekin~D Cubuk, Alex Kurakin, Han Zhang, and Colin Raffel.
\newblock Fixmatch: Simplifying semi-supervised learning with consistency and
  confidence.
\newblock {\em arXiv preprint arXiv:2001.07685}, 2020.

\bibitem{lee2013pseudo}
Dong-Hyun Lee.
\newblock Pseudo-label: The simple and efficient semi-supervised learning
  method for deep neural networks.
\newblock In {\em Workshop on challenges in representation learning, ICML},
  volume~3, page~2, 2013.

\bibitem{xie2019unsupervised}
Qizhe Xie, Zihang Dai, Eduard Hovy, Minh-Thang Luong, and Quoc~V Le.
\newblock Unsupervised data augmentation for consistency training.
\newblock 2019.

\bibitem{xie2019self}
Qizhe Xie, Eduard Hovy, Minh-Thang Luong, and Quoc~V Le.
\newblock Self-training with noisy student improves imagenet classification.
\newblock {\em arXiv preprint arXiv:1911.04252}, 2019.

\bibitem{blundell2015weight}
Charles Blundell, Julien Cornebise, Koray Kavukcuoglu, and Daan Wierstra.
\newblock Weight uncertainty in neural networks.
\newblock {\em arXiv preprint arXiv:1505.05424}, 2015.

\bibitem{gal2016dropout}
Yarin Gal and Zoubin Ghahramani.
\newblock Dropout as a bayesian approximation: Representing model uncertainty
  in deep learning.
\newblock In {\em international conference on machine learning}, pages
  1050--1059, 2016.

\bibitem{maddox2019simple}
Wesley Maddox, Timur Garipov, Pavel Izmailov, Dmitry Vetrov, and Andrew~Gordon
  Wilson.
\newblock A simple baseline for {B}ayesian uncertainty in deep learning.
\newblock {\em arXiv preprint arXiv:1902.02476}, 2019.

\bibitem{franchi2019tradi}
Gianni Franchi, Andrei Bursuc, Emanuel Aldea, S{\'e}verine Dubuisson, and
  Isabelle Bloch.
\newblock {TRADI}: Tracking deep neural network weight distributions.
\newblock {\em arXiv preprint arXiv:1912.11316}, 2019.

\bibitem{lakshminarayanan2017simple}
Balaji Lakshminarayanan, Alexander Pritzel, and Charles Blundell.
\newblock Simple and scalable predictive uncertainty estimation using deep
  ensembles.
\newblock In {\em Advances in Neural Information Processing Systems}, pages
  6402--6413, 2017.

\bibitem{liang2017enhancing}
Shiyu Liang, Yixuan Li, and Rayadurgam Srikant.
\newblock Enhancing the reliability of out-of-distribution image detection in
  neural networks.
\newblock {\em arXiv preprint arXiv:1706.02690}, 2017.

\bibitem{Khosla_Stanford_Dogs_FGVC2011}
Aditya Khosla, Nityananda Jayadevaprakash, Bangpeng Yao, and Li~Fei-Fei.
\newblock Novel dataset for fine-grained image categorization.
\newblock In {\em First Workshop on Fine-Grained Visual Categorization, IEEE
  Conference on Computer Vision and Pattern Recognition (CVPR)}, June 2011.

\bibitem{parkhi_oxford_pets_2012}
O.~M. Parkhi, A.~Vedaldi, A.~Zisserman, and C.~V. Jawahar.
\newblock Cats and dogs.
\newblock In {\em IEEE Conference on Computer Vision and Pattern Recognition
  (CVPR)}, 2012.

\bibitem{WahCUB_200_2011}
C.~Wah, S.~Branson, P.~Welinder, P.~Perona, and S.~Belongie.
\newblock The caltech-ucsd birds-200-2011 dataset.
\newblock Technical Report CNS-TR-2011-001, California Institute of Technology,
  2011.

\bibitem{VanHorn_nabirds_2015}
G.~{Van Horn}, S.~{Branson}, R.~{Farrell}, S.~{Haber}, J.~{Barry},
  P.~{Ipeirotis}, P.~{Perona}, and S.~{Belongie}.
\newblock Building a bird recognition app and large scale dataset with citizen
  scientists: The fine print in fine-grained dataset collection.
\newblock In {\em IEEE Conference on Computer Vision and Pattern Recognition
  (CVPR)}, pages 595--604, June 2015.

\bibitem{Krause_Stanford_Cars_2013}
Jonathan Krause, Michael Stark, Jia Deng, and Li~Fei-Fei.
\newblock 3d object representations for fine-grained categorization.
\newblock In {\em International IEEE Workshop on 3D Representation and
  Recognition (3dRR-13)}, 2013.

\bibitem{Yang_cars_2015}
L.~{Yang}, P.~{Luo}, C.~C. {Loy}, and X.~{Tang}.
\newblock A large-scale car dataset for fine-grained categorization and
  verification.
\newblock In {\em IEEE Conference on Computer Vision and Pattern Recognition
  (CVPR)}, pages 3973--3981, June 2015.

\bibitem{Maji_airplanes_2013}
S.~Maji, J.~Kannala, E.~Rahtu, M.~Blaschko, and A.~Vedaldi.
\newblock Fine-grained visual classification of aircraft.
\newblock Technical report, arXiv:1306.5151, 2013.

\bibitem{Nilsback_flowers_2008}
M.~{Nilsback} and A.~{Zisserman}.
\newblock Automated flower classification over a large number of classes.
\newblock In {\em Indian Conference on Computer Vision, Graphics Image
  Processing}, pages 722--729, Dec 2008.

\bibitem{Horn_iNat_2018}
G.~V. {Horn}, O.~M. {Aodha}, Y.~{Song}, Y.~{Cui}, C.~{Sun}, A.~{Shepard},
  H.~{Adam}, P.~{Perona}, and S.~{Belongie}.
\newblock The inaturalist species classification and detection dataset.
\newblock In {\em IEEE Conference on Computer Vision and Pattern Recognition
  (CVPR}, pages 8769--8778, June 2018.

\bibitem{Mao_1993_nonlinear_analysis}
J.~{Mao} and A.~K. {Jain}.
\newblock Discriminant analysis neural networks.
\newblock In {\em International Conference on Neural Networks}, pages 300--305
  vol.1, 1993.

\bibitem{Wang_ijcai2017_2D_LDA}
Qi~Wang, Zequn Qin, Feiping Nie, and Yuan Yuan.
\newblock Convolutional 2d lda for nonlinear dimensionality reduction.
\newblock In {\em International Joint Conference on Artificial Intelligence
  (IJCAI)}, pages 2929--2935, 2017.

\bibitem{Dorfer_ICLR16_deep_LDA}
Matthias Dorfer, Rainer Kelz, and Gerhard Widmer.
\newblock Deep linear discriminant analysis.
\newblock In {\em International Conference on Learning Representations (ICLR)},
  pages 1--11, 2016.

\bibitem{Zhong_CDA_ICPR_2018}
G.~{Zhong}, Y.~{Zheng}, X.~{Zhang}, H.~{Wei}, and X.~{Ling}.
\newblock Convolutional discriminant analysis.
\newblock In {\em International Conference on Pattern Recognition (ICPR)},
  pages 1456--1461, 2018.

\bibitem{Li_ciss19_da}
L.~{Li}, M.~{Doroslovački}, and M.H. {Loew}.
\newblock Discriminant analysis deep neural networks.
\newblock In {\em Conference on Information Sciences and Systems (CISS)}, pages
  1--6, 2019.

\bibitem{annomynous_nda}
Anonymous.
\newblock Neural discriminant analysis for fine-grained classification.
\newblock In {\em ICIP}, 2020.
\newblock accepted, see supplemental material.

\bibitem{dvornik2019diversity}
Nikita Dvornik, Cordelia Schmid, and Julien Mairal.
\newblock Diversity with cooperation: Ensemble methods for few-shot
  classification.
\newblock In {\em Proceedings of the IEEE International Conference on Computer
  Vision}, pages 3723--3731, 2019.

\bibitem{gidaris2019boosting}
Spyros Gidaris, Andrei Bursuc, Nikos Komodakis, Patrick P{\'e}rez, and Matthieu
  Cord.
\newblock Boosting few-shot visual learning with self-supervision.
\newblock In {\em Proceedings of the IEEE International Conference on Computer
  Vision}, pages 8059--8068, 2019.

\bibitem{franchi2020OVNNI}
Gianni Franchi, Andrei Bursuc, Emanuel Aldea, S{\'e}verine Dubuisson, and
  Isabelle Bloch.
\newblock One versus all for deep neural network incertitude {(OVNNI)}
  quantification.
\newblock {\em arXiv preprint arXiv:2006.00954}, 2019.

\bibitem{corbiere2019addressing}
Charles Corbi{\`e}re, Nicolas Thome, Avner Bar-Hen, Matthieu Cord, and Patrick
  P{\'e}rez.
\newblock Addressing failure prediction by learning model confidence.
\newblock In {\em Advances in Neural Information Processing Systems}, pages
  2898--2909, 2019.

\bibitem{Farrell_2011}
R.~{Farrell}, O.~{Oza}, {Ning Zhang}, V.~I. {Morariu}, T.~{Darrell}, and L.~S.
  {Davis}.
\newblock Birdlets: Subordinate categorization using volumetric primitives and
  pose-normalized appearance.
\newblock In {\em International Conference on Computer Vision (ICCV)}, pages
  161--168, Nov 2011.

\bibitem{Parkhi_cats_dogs_2011}
O.~M. {Parkhi}, A.~{Vedaldi}, C.~V. {Jawahar}, and A.~{Zisserman}.
\newblock The truth about cats and dogs.
\newblock In {\em International Conference on Computer Vision (ICCV)}, pages
  1427--1434, Nov 2011.

\bibitem{Liu_dog_part_2012}
Jiongxin Liu, Angjoo Kanazawa, David Jacobs, and Peter Belhumeur.
\newblock Dog breed classification using part localization.
\newblock In {\em European Conference on Computer Vision (ECCV)}, pages
  172--185, 2012.

\bibitem{branson2014bird}
Steve Branson, Grant Van~Horn, Serge Belongie, and Pietro Perona.
\newblock Bird species categorization using pose normalized deep convolutional
  nets.
\newblock In {\em British Machine Vision Conference (BMVC)}, 2014.

\bibitem{ZhangECCV14}
Ning Zhang, Jeff Donahue, Ross Girshick, and Trevor Darrell.
\newblock Part-based r-cnn for fine-grained category detection.
\newblock In {\em European Conference on Computer Vision (ECCV)}, 2014.

\bibitem{ZhangCVPR14}
Ning Zhang, Manohar Paluri, Marc'Aurelio Rantazo, Trevor Darrell, and Lubomir
  Bourdev.
\newblock Panda: Pose aligned networks for deep attribute modeling.
\newblock In {\em IEEE Conference on Computer Vision and Pattern Recognition
  (CVPR)}, 2014.

\bibitem{Krause_parts_2014}
J.~{Krause}, T.~{Gebru}, J.~{Deng}, L.~{Li}, and L.~{Fei-Fei}.
\newblock Learning features and parts for fine-grained recognition.
\newblock In {\em International Conference on Pattern Recognition (ICPR)},
  pages 26--33, Aug 2014.

\bibitem{Lin_deep_LAC_2015}
D.~{Lin}, X.~{Shen}, C.~{Lu}, and J.~{Jia}.
\newblock Deep lac: Deep localization, alignment and classification for
  fine-grained recognition.
\newblock In {\em Conference on Computer Vision and Pattern Recognition
  (CVPR)}, pages 1666--1674, June 2015.

\bibitem{ZhangSGD15}
Ning Zhang, Evan Shelhamer, Yang Gao, and Trevor Darrell.
\newblock Fine-grained pose prediction, normalization, and recognition.
\newblock {\em CoRR}, abs/1511.07063, 2015.

\bibitem{Huang_2016_CVPR}
Shaoli Huang, Zhe Xu, Dacheng Tao, and Ya~Zhang.
\newblock Part-stacked cnn for fine-grained visual categorization.
\newblock In {\em IEEE Conference on Computer Vision and Pattern Recognition
  (CVPR)}, June 2016.

\bibitem{Zhang_SPDA_CNN_2016}
H.~{Zhang}, T.~{Xu}, M.~{Elhoseiny}, X.~{Huang}, S.~{Zhang}, A.~{Elgammal}, and
  D.~{Metaxas}.
\newblock Spda-cnn: Unifying semantic part detection and abstraction for
  fine-grained recognition.
\newblock In {\em IEEE Conference on Computer Vision and Pattern Recognition
  (CVPR)}, pages 1143--1152, June 2016.

\bibitem{Vedaldi_detail_attribute_CVPR_2014}
Andrea Vedaldi, Siddharth Mahendran, Stavros Tsogkas, Subhransu Maji, Ross~B.
  Girshick, Juho Kannala, Esa Rahtu, Iasonas Kokkinos, Matthew~B. Blaschko,
  David~J. Weiss, Ben Taskar, Karen Simonyan, Naomi Saphra, and Sammy Mohamed.
\newblock Understanding objects in detail with fine-grained attributes.
\newblock {\em IEEE Conference on Computer Vision and Pattern Recognition},
  pages 3622--3629, 2014.

\bibitem{Reed_visual_descriptions_CVPR_2016}
Scott~E. Reed, Zeynep Akata, Honglak Lee, and Bernt Schiele.
\newblock Learning deep representations of fine-grained visual descriptions.
\newblock In {\em {CVPR}}, pages 49--58, 2016.

\bibitem{He_vision_language_CVPR_2017}
Xiangteng He and Yuxin Peng.
\newblock Fine-grained image classification via combining vision and language.
\newblock In {\em {CVPR}}, pages 7332--7340, 2017.

\bibitem{Xu_semantic_embedding_IJCAI_2018}
Huapeng Xu, Guilin Qi, Jingjing Li, Meng Wang, Kang Xu, and Huan Gao.
\newblock Fine-grained image classification by visual-semantic embedding.
\newblock In {\em Proceedings of the International Joint Conference on
  Artificial Intelligence (IJCAI)}, pages 1043--1049, 7 2018.

\bibitem{Xiao_attention_2015}
Tianjun Xiao, Yichong Xu, Kuiyuan Yang, Jiaxing Zhang, Yuxin Peng, and
  Z.~Zhang.
\newblock The application of two-level attention models in deep convolutional
  neural network for fine-grained image classification.
\newblock In {\em 2015 IEEE Conference on Computer Vision and Pattern
  Recognition (CVPR)}, pages 842--850, jun 2015.

\bibitem{Liu_FullyCA_2016}
Xiao Liu, Tian Xia, Jiang Wang, and Yuanqing Lin.
\newblock Fully convolutional attention localization networks: Efficient
  attention localization for fine-grained recognition.
\newblock {\em CoRR}, abs/1603.06765, 2016.

\bibitem{Zheng_multi_attention_2017}
H.~{Zheng}, J.~{Fu}, T.~{Mei}, and J.~{Luo}.
\newblock Learning multi-attention convolutional neural network for
  fine-grained image recognition.
\newblock In {\em IEEE International Conference on Computer Vision (ICCV)},
  pages 5219--5227, Oct 2017.

\bibitem{Zhao_attention_2017}
Bo~Zhao, Xiao Wu, Jiashi Feng, Qiang Peng, and Shuicheng Yan.
\newblock Diversified visual attention networks for fine-grained object
  classification.
\newblock {\em IEEE Transactions on Multimedia}, 19(6):1245--1256, June 2017.

\bibitem{Fu_look_closer_2017}
J.~{Fu}, H.~{Zheng}, and T.~{Mei}.
\newblock Look closer to see better: Recurrent attention convolutional neural
  network for fine-grained image recognition.
\newblock In {\em IEEE Conference on Computer Vision and Pattern Recognition
  (CVPR)}, pages 4476--4484, July 2017.

\bibitem{Peng_object_part_attention_TIP_2018}
Yuxin Peng, Xiangteng He, and Junjie Zhao.
\newblock Object-part attention model for fine-grained image classification.
\newblock {\em {IEEE} Trans. Image Processing (TIP)}, 27(3):1487--1500, 2018.

\bibitem{Yang_learn_navigate_2018}
Ze~Yang, Tiange Luo, Dong Wang, Zhiqiang Hu, Jun Gao, and Liwei Wang.
\newblock Learning to navigate for fine-grained classification.
\newblock In {\em European Conference on Computer Vision (ECCV)}, September
  2018.

\bibitem{Sun_multi_attention_2018}
Ming Sun, Yuchen Yuan, Feng Zhou, and Errui Ding.
\newblock Multi-attention multi-class constraint for fine-grained image
  recognition.
\newblock In {\em European Conference on Computer Vision (ECCV)}, September
  2018.

\bibitem{Wang_granular_descriptor_2015}
D.~{Wang}, Z.~{Shen}, J.~{Shao}, W.~{Zhang}, X.~{Xue}, and Z.~{Zhang}.
\newblock Multiple granularity descriptors for fine-grained categorization.
\newblock In {\em IEEE International Conference on Computer Vision (ICCV)},
  pages 2399--2406, Dec 2015.

\bibitem{Zhang_picking_resposes_2016}
X.~{Zhang}, H.~{Xiong}, W.~{Zhou}, W.~{Lin}, and Q.~{Tian}.
\newblock Picking deep filter responses for fine-grained image recognition.
\newblock In {\em IEEE Conference on Computer Vision and Pattern Recognition
  (CVPR)}, pages 1134--1142, June 2016.

\bibitem{Lin_bilinear_pooling_2015}
T.~{Lin}, A.~{RoyChowdhury}, and S.~{Maji}.
\newblock Bilinear cnn models for fine-grained visual recognition.
\newblock In {\em IEEE International Conference on Computer Vision (ICCV)},
  pages 1449--1457, Dec 2015.

\bibitem{Gao_compact_pooling_2016}
Y.~{Gao}, O.~{Beijbom}, N.~{Zhang}, and T.~{Darrell}.
\newblock Compact bilinear pooling.
\newblock In {\em IEEE Conference on Computer Vision and Pattern Recognition
  (CVPR)}, pages 317--326, June 2016.

\bibitem{Lin_improve_pooling_2017}
Tsung-Yu Lin and Subhransu Maji.
\newblock Improved bilinear pooling with cnns.
\newblock In {\em Proceedings of the British Machine Vision Conference (BMVC)},
  pages 117.1--117.12, September 2017.

\bibitem{Cui_kernel_pooling_2017}
Y.~{Cui}, F.~{Zhou}, J.~{Wang}, X.~{Liu}, Y.~{Lin}, and S.~{Belongie}.
\newblock Kernel pooling for convolutional neural networks.
\newblock In {\em IEEE Conference on Computer Vision and Pattern Recognition
  (CVPR)}, pages 3049--3058, July 2017.

\bibitem{Cai_high_order_hierarchical_2017}
S.~{Cai}, W.~{Zuo}, and L.~{Zhang}.
\newblock Higher-order integration of hierarchical convolutional activations
  for fine-grained visual categorization.
\newblock In {\em IEEE International Conference on Computer Vision (ICCV)},
  pages 511--520, Oct 2017.

\bibitem{Yu_hierarchical_pooling_2018}
Chaojian Yu, Xinyi Zhao, Qi~Zheng, Peng Zhang, and Xinge You.
\newblock Hierarchical bilinear pooling for fine-grained visual recognition.
\newblock In {\em European Conference on Computer Vision (ECCV)}, September
  2018.

\bibitem{Wei_Grassmann_pooling_2018}
Xing Wei, Yue Zhang, Yihong Gong, Jiawei Zhang, and Nanning Zheng.
\newblock Grassmann pooling as compact homogeneous bilinear pooling for
  fine-grained visual classification.
\newblock In {\em The European Conference on Computer Vision (ECCV)}, September
  2018.

\bibitem{Xie_myperclass_augment_CVPR_2015}
S.~{Xie}, T.~{Yang}, {Xiaoyu Wang}, and {Yuanqing Lin}.
\newblock Hyper-class augmented and regularized deep learning for fine-grained
  image classification.
\newblock In {\em IEEE Conference on Computer Vision and Pattern Recognition
  (CVPR)}, pages 2645--2654, June 2015.

\bibitem{Krause_noisy_data_ECCV_2016}
Jonathan Krause, Benjamin Sapp, Andrew Howard, Howard Zhou, Alexander Toshev,
  Tom Duerig, James Philbin, and Li~Fei{-}Fei.
\newblock The unreasonable effectiveness of noisy data for fine-grained
  recognition.
\newblock In {\em {ECCV}}, volume 9907, pages 301--320, 2016.

\bibitem{Gebru_in_the_wild_ICCV_2017}
Timnit Gebru, Judy Hoffman, and Li~Fei-Fei.
\newblock Fine-grained recognition in the wild: A multi-task domain adaptation
  approach.
\newblock {\em IEEE International Conference on Computer Vision (ICCV)}, pages
  1358--1367, 2017.

\bibitem{Zhang_auxiliary_data_ECCV_2018}
Yabin Zhang, Hui Tang, and Kui Jia.
\newblock Fine-grained visual categorization using meta-learning optimization
  with sample selection of auxiliary data.
\newblock In {\em European Conference on Computer Vision (ECCV)}, September
  2018.

\bibitem{Xu_webly_supervised_PAMI_2018}
Z.~{Xu}, S.~{Huang}, Y.~{Zhang}, and D.~{Tao}.
\newblock Webly-supervised fine-grained visual categorization via deep domain
  adaptation.
\newblock {\em IEEE Transactions on Pattern Analysis and Machine Intelligence
  (PAMI)}, 40(5):1100--1113, May 2018.

\bibitem{Cui_iNatTransfer_CVPR_2018}
Chen Sun Andrew Howard Serge~Belongie Yin~Cui, Yang~Song.
\newblock Large scale fine-grained categorization and domain-specific transfer
  learning.
\newblock In {\em CVPR}, 2018.

\bibitem{Branson_human_in_loop_ECCV_2010}
Steve Branson, Catherine Wah, Florian Schroff, Boris Babenko, Peter Welinder,
  Pietro Perona, and Serge Belongie.
\newblock Visual recognition with humans in the loop.
\newblock In {\em European Conference on Computer Vision (ECCV)}, pages
  438--451, 2010.

\bibitem{Cui_bootstrap_CVPR_2016}
Yin Cui, Feng Zhou, Yuanqing Lin, and Serge Belongie.
\newblock Fine-grained categorization and dataset bootstrapping using deep
  metric learning with humans in the loop.
\newblock In {\em Computer Vision and Pattern Recognition (CVPR)}, 2016.

\bibitem{Deng_crowd_wisdom_PAMI_2016}
J.~{Deng}, J.~{Krause}, M.~{Stark}, and L.~{Fei-Fei}.
\newblock Leveraging the wisdom of the crowd for fine-grained recognition.
\newblock {\em IEEE Transactions on Pattern Analysis and Machine Intelligence
  (PAMI)}, 38(4):666--676, April 2016.

\bibitem{Szegedy_Rethinking_Inception_CVPR_2016}
Christian Szegedy, Vincent Vanhoucke, Sergey Ioffe, Jonathon Shlens, and
  Zbigniew Wojna.
\newblock Rethinking the inception architecture for computer vision.
\newblock {\em IEEE Conference on Computer Vision and Pattern Recognition
  (CVPR)}, pages 2818--2826, 2016.

\bibitem{He_ResNet_CVPR_2016}
K.~{He}, X.~{Zhang}, S.~{Ren}, and J.~{Sun}.
\newblock Deep residual learning for image recognition.
\newblock In {\em IEEE Conference on Computer Vision and Pattern Recognition
  (CVPR)}, pages 770--778, June 2016.

\bibitem{He_ResNet_ECCV_2016}
Kaiming He, Xiangyu Zhang, Shaoqing Ren, and Jian Sun.
\newblock Identity mappings in deep residual networks.
\newblock In {\em ECCV}, 2016.

\bibitem{snell2017prototypical}
Jake Snell, Kevin Swersky, and Richard Zemel.
\newblock Prototypical networks for few-shot learning.
\newblock In {\em Advances in neural information processing systems}, pages
  4077--4087, 2017.

\bibitem{krizhevsky2009learning}
Alex Krizhevsky, Geoffrey Hinton, et~al.
\newblock Learning multiple layers of features from tiny images.
\newblock 2009.

\bibitem{Krizhevsky_NIPS2012_AlexNet}
Alex Krizhevsky, Ilya Sutskever, and Geoffrey~E Hinton.
\newblock Imagenet classification with deep convolutional neural networks.
\newblock pages 1097--1105, 2012.

\bibitem{zagoruyko2016wide}
Sergey Zagoruyko and Nikos Komodakis.
\newblock Wide residual networks.
\newblock {\em arXiv preprint arXiv:1605.07146}, 2016.

\bibitem{Cubuk2019_RandAugment}
Ekin~D Cubuk, Barret Zoph, Jonathon Shlens, and Quoc~V Le.
\newblock {RandAugment}: Practical automated data augmentation with a reduced
  search space.
\newblock {\em arXiv preprint arXiv:1909.13719}, 2019.

\bibitem{hendrycks2016baseline}
Dan Hendrycks and Kevin Gimpel.
\newblock A baseline for detecting misclassified and out-of-distribution
  examples in neural networks.
\newblock {\em arXiv preprint arXiv:1610.02136}, 2016.

\bibitem{guo2017calibration}
Chuan Guo, Geoff Pleiss, Yu~Sun, and Kilian~Q Weinberger.
\newblock On calibration of modern neural networks.
\newblock In {\em Proceedings of the 34th International Conference on Machine
  Learning-Volume 70}, pages 1321--1330. JMLR. org, 2017.

\end{thebibliography}
\end{document}